\documentclass[letterpaper, 10 pt, journal, twoside]{IEEEtran}

\IEEEoverridecommandlockouts

\usepackage{amsmath,amsfonts,amssymb}
\usepackage{algpseudocode}
\usepackage{algorithm}
\usepackage{array}
\usepackage{svg}
\usepackage{textcomp}
\usepackage{stfloats}
\usepackage{url}
\usepackage{verbatim}
\usepackage{graphicx}
\usepackage{cite}
\usepackage{wrapfig}
\usepackage{multirow}
\usepackage{microtype}
\usepackage{tabularx,ragged2e}
\usepackage{booktabs}
\usepackage[letterpaper, left=0.667in, right=0.667in, bottom=0.597in, top=0.792in]{geometry}
\usepackage{fancyhdr}

\usepackage{graphicx}
\usepackage{subcaption}
\usepackage{threeparttable}
\usepackage[normalem]{ulem}   

\makeatletter
\let\NAT@parse\undefined
\makeatother
\usepackage{hyperref}

\begin{document}

\UseRawInputEncoding

\title{Dynamic Policy Learning for Legged Robot with Simplified Model Pretraining and Model-Homotopy-Inspired Transfer}

\markboth{IEEE Robotics and Automation Letters. Preprint Version. Accepted April, 2026}
{Kang \MakeLowercase{\textit{et al.}}: Dynamic Policy Learning for Legged Robot with Simplified Model Pretraining and Model-Homotopy-Inspired Transfer} 

\author{Dongyun Kang$^{1}$, Min-Gyu Kim$^{1}$, Tae-Gyu Song$^{1}$, Hajun Kim$^{1}$, Sehoon Ha$^{2}$, and Hae-Won Park$^{1}$  
\thanks{Manuscript received: December, 30, 2025; Revised March, 24, 2026; Accepted April, 18, 2026.} 
\thanks{This paper was recommended for publication by Editor Wei Pan upon evaluation of the Associate Editor and Reviewers' comments.
This work was supported by the Technology Innovation Program(or Industrial Strategic Technology Development Program-Robot Industry Technology Development)(RS-2024-00427719, Dexterous and Agile Humanoid Robots for Industrial Applications) funded by the Ministry of Trade Industry \& Energy(MOTIE, Korea)
} 
\thanks{$^{1}$D. Kang, M.-G. Kim, T.-G. Song, H. Kim, and H.-W. Park are with the Department of Mechanical Engineering, Korea Advanced Institute of Science and Technology (KAIST), Daejeon 34141, Republic of Korea (e-mail: {\tt\footnotesize haewonpark@kaist.ac.kr}).}%
\thanks{$^{2}$S. Ha is with the School of Interactive Computing, Georgia Institute of Technology, Atlanta, GA 30308, USA (e-mail: {\tt\footnotesize sehoonha@gatech.edu}).}%
\thanks{Digital Object Identifier (DOI): see top of this page.} 
}

\maketitle

\begin{abstract}
Generating dynamic motions for legged robots remains a challenging problem. While reinforcement learning has achieved notable success in various legged locomotion tasks, producing highly dynamic behaviors often requires extensive reward tuning or high-quality demonstrations. 
Leveraging reduced-order models can help mitigate these challenges. However, the model discrepancy poses a significant challenge when transferring policies to full-body dynamics environments.
In this work, we introduce a continuation-based learning framework that combines simplified model pretraining and model-homotopy-inspired transfer to efficiently generate and refine complex dynamic behaviors. First, we pretrain the policy using a single rigid body model to capture core motion patterns in a simplified environment. 
Next, we employ a continuation strategy to progressively transfer the policy to the full-body environment, minimizing performance loss. 
To define the continuation path, we introduce a parametric transition path from the single rigid body model to the full-body model by gradually redistributing mass and inertia between the trunk and legs.
The proposed method achieves faster convergence and demonstrates superior stability during the transfer process compared to baseline methods. Our framework is validated on a range of dynamic tasks, including flips and wall-assisted maneuvers, and is successfully deployed on a real quadrupedal robot.
\end{abstract}


\section{Introduction}
\label{sec:Introduction}

\IEEEPARstart{L}{earning} complex and dynamic motions for legged robots remains a challenging problem. Successfully generating such behaviors requires navigating the intricate interplay between the robot's body momentum, leg inertia, and contact forces, making it difficult to develop effective control strategies.


Classical motion generation approaches, particularly trajectory optimization (TO), formulate the problem as a constrained optimization\cite{SynthesisAndStabilizationOfComplexBehaviors2012,DiscoveryofComplexBehaviors2012,MultiContactLocomotionCarpentier2018,ContactImplicitMPC2023,
LIPTO2018,MitCheetah3MPC2018,TOWR2018,NMPC2020,BezierSRBTO2025,WBIC2019,CentroidalDynamicsAndFullKinematics2014,MomentumAwareTO2022,StagedContactOptimization2023}. While TO offers interpretability and principled motion synthesis, solving it reliably is often hindered by nonconvexity and the hybrid nature of contact dynamics. 
To cope with these challenges, reduced-order models are employed to approximate feasible motions\cite{LIPTO2018,MitCheetah3MPC2018,TOWR2018,NMPC2020,BezierSRBTO2025,WBIC2019,CentroidalDynamicsAndFullKinematics2014,MomentumAwareTO2022,StagedContactOptimization2023}, yet relying on them introduces dynamics mismatches. 
While hierarchical frameworks\cite{WBIC2019,MomentumAwareTO2022,StagedContactOptimization2023,ContactTimingandTOfor3dJumping2022} mitigate this by refining trajectories with full-body dynamics, this process is often computationally intensive, making it challenging to achieve real-time reactivity without significant optimization expertise.

\begin{figure}
    \centering
    \includegraphics[width=1.0\columnwidth]{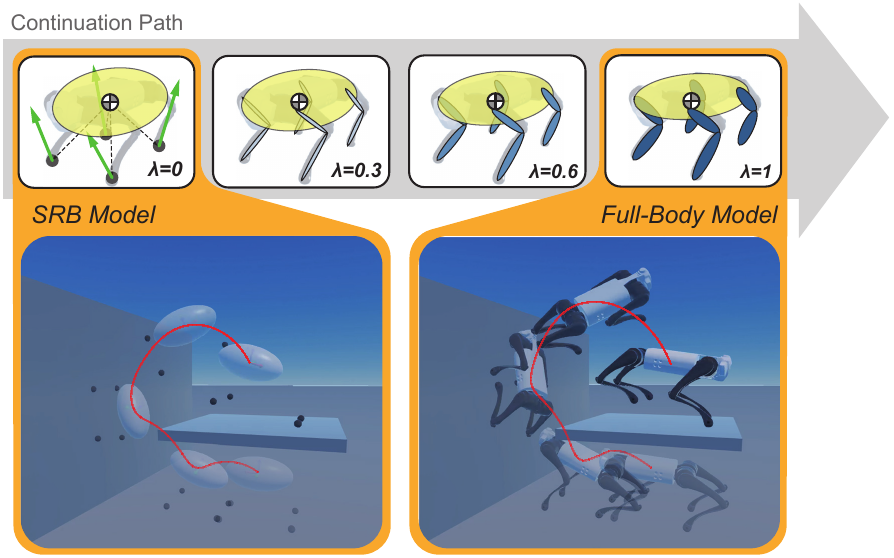}
    \captionsetup{font=small}
    \caption{
    Conceptual illustration of the proposed approach. The framework defines a continuation path from the simplified SRB to full-body dynamics via gradual mass and inertia redistribution. This approach facilitates the stable transfer of core motion patterns to complex, highly dynamic behaviors such as wall-assisted maneuvers.
    }
    \label{figure:model_homotopic_path}
    \vspace{-0.4cm}
\end{figure}

Reinforcement learning (RL), which has demonstrated impressive results across various legged locomotion tasks
\cite{ChallengingTerrain2020, AnymalParkour2024,ComplexTerrainNavigation2025}, offers an alternative paradigm capable of discovering highly dynamic behaviors. 
However, generating complex motions without prior information often results in instability and convergence to suboptimal solutions. Addressing these issues requires extensive reward engineering, making the process time consuming and difficult to apply to a wide range of dynamic behaviors. While imitation learning\cite{RLDemonstrationsFromTO2022,ImitatingAnimalLocomotion2020,DeepMimic2018,AMP2021}, which utilizes expert demonstrations, is an effective solution for guiding RL, obtaining high-quality demonstrations often requires substantial resources and is sometimes not readily available\cite{Wasabi2023,LearningSpringMassLocomotion2021, OptimizingBipedalManeuversSRB2022, OptMimic2023}.

One strategy to alleviate these challenges is to leverage a reduced-order model to train the policy in a simpler environment\cite{GLiDE2022, Brachiation2022, RLforROM2024, AdaptiveTrackingofSRBCharacter2023}. This approach offers several advantages, such as simple reward design and faster simulation speeds, leading to more efficient policy learning\cite{GLiDE2022,Brachiation2022}.
However, the discrepancy between simplified models and full-body models, commonly referred to as the model gap, poses a significant hurdle when deploying these policies in real environments. 

Previous works have addressed this gap by combining simplified model planners with whole-body tracking controllers\cite{GLiDE2022,RLforROM2024} or employing imitation learning \cite{Brachiation2022}. Although these methods have shown promise in simple locomotion tasks or 2D scenarios, their applicability to highly dynamic maneuvers remains unproven.
Moreover, adhering to the simplified model's motion often leads to suboptimal solutions for the full-body system. 
These observations highlight the need for an effective and stable refinement process to bridge the model gap.

Continuation methods~\cite{NumericalContinuation2012} offer a powerful conceptual framework to tackle such challenges. By gradually transforming a well-understood, simpler system into a more complex target system, continuation methods facilitate smoother transitions and more robust convergence on difficult problems. 
Building on this conceptual foundation, we propose a continuation-based method that first leverages a simplified model to pretrain the policy in an idealized setting, and then incrementally bridges the gap to the full-body environment. 
To define the continuation path, we introduce a parametric transition path from the Single Rigid Body (SRB) model to the full-body model, constructed by progressively interpolating the mass and inertia between the trunk and legs.
Following this controlled progression allows the policy to navigate increasingly complex dynamics and achieve a smooth, reliable transfer, as conceptually illustrated in Fig. 1.

We validate our framework on a series of challenging dynamic tasks, including flips and wall-assisted maneuvers. Our results show that the proposed approach enhances learning stability and achieves faster convergence compared to baseline methods. Finally, we successfully transfer the learned policies to a real quadrupedal robot.

In summary, the key contributions of this work are as follows:

\begin{itemize} 
\item We present a continuation-based learning framework that integrates simplified model pretraining and model-homotopy-inspired transfer to efficiently generate and refine complex dynamic motions. 
\item This framework provides a principled way to smoothly transition from simplified to full-body dynamics, improving both convergence speed and learning stability.
\item We validate the effectiveness of our method by successfully generating a variety of dynamic behaviors in simulation, and deploy the learned policies on a real robot. 
\end{itemize}


\section{Related Work}
\label{sec:RelatedWork}

\subsection{Trajectory Optimization}

TO is a systematic process for finding the optimal trajectory that satisfies a set of physical and task constraints. Due to its interpretability and ability to handle diverse constraints, TO has been extensively utilized in legged robot motion generation \cite{SynthesisAndStabilizationOfComplexBehaviors2012,DiscoveryofComplexBehaviors2012,MultiContactLocomotionCarpentier2018,ContactImplicitMPC2023}. However, the optimization landscape for legged locomotion is fundamentally nonconvex. Furthermore, the hybrid nature of the system, characterized by discontinuous contact modes, makes finding desired solutions extremely challenging, as solvers are prone to getting trapped in poor local optima.

To overcome these computational complexities, a common strategy is to decompose the high-dimensional full-body optimization problem into simpler, manageable subproblems. This is typically achieved by pre-specifying contact sequences\cite{MitCheetah3MPC2018,NMPC2020,MomentumAwareTO2022} or employing reduced-order models, such as Linear Inverted Pendulum (LIP)\cite{LIPTO2018}, SRB\cite{MitCheetah3MPC2018,TOWR2018,NMPC2020,BezierSRBTO2025,WBIC2019}, and centroidal model\cite{CentroidalDynamicsAndFullKinematics2014,MomentumAwareTO2022,StagedContactOptimization2023}, to generate approximate solutions. 
These simplified solutions are often used as warm starts or references for hierarchical optimization frameworks, which subsequently refine the trajectory using a detailed full-body model to ensure dynamic feasibility and tracking accuracy \cite{WBIC2019,MomentumAwareTO2022,ContactTimingandTOfor3dJumping2022,StagedContactOptimization2023}.


\subsection{Reinforcement Learning}
An alternative approach for achieving dynamic motions is RL. RL has recently achieved remarkable success across various tasks, including quadruped locomotion \cite{ChallengingTerrain2020, AnymalParkour2024,ComplexTerrainNavigation2025}. 

However, generating complex, high-quality motions without prior information often requires tedious reward tuning.
To guide the learning more effectively, several strategies have been proposed, such as curriculum learning \cite{SymmetricLocomotion2018, RobotParkourLearning2023,ComplexTerrainNavigation2025}, the use of barrier rewards to enforce constraints \cite{NotOnlyRewardsButAlsoConstraints2024, BarrierBasedStyleReward2024}, and combining RL with model-based control \cite{IFM2023}. Another effective approach to address the reward engineering problem is to leverage expert demonstrations. These demonstrations can be obtained through hand-designed motions, trajectory optimization \cite{RLDemonstrationsFromTO2022}, or motion capture data \cite{ImitatingAnimalLocomotion2020}. Once these demonstrations are available, imitation can be performed using imitation rewards \cite{DeepMimic2018} or adversarial rewards to capture stylistic behaviors \cite{AMP2021}.
However, collecting high-quality demonstrations often requires significant resources. To mitigate this issue, Li et al. \cite{Wasabi2023} proposed a method that utilizes rough partial demonstrations obtained from hand-held human demonstrations. Additionally, trajectory optimization on simplified models has been used as a more accessible and less resource-intensive way to generate reasonable demonstrations \cite{LearningSpringMassLocomotion2021, OptimizingBipedalManeuversSRB2022, OptMimic2023}.

\subsection{Policy Learning with Simplified Model}

A few notable studies \cite{GLiDE2022, Brachiation2022, RLforROM2024, AdaptiveTrackingofSRBCharacter2023} have focused on utilizing RL policies learned in environments governed by simplified dynamics models. For example, Xie et al. \cite{GLiDE2022} used a policy trained in a SRB environment, where the RL policy generated desired body accelerations that were tracked by a quadratic programming-based tracking controller. This study highlighted the benefits of simplified model learning, such as simple reward design, fast and efficient training simulations, and robust sim-to-real deployment. Reda et al. \cite{Brachiation2022} used a point mass with arm model to train a policy, which served as a motion planner for the full-body model and was subsequently refined through imitation learning. Chen et al. \cite{RLforROM2024}, employed a LIP model, and the resulting policy were tracked using an operational space control in the full-body environment.

These works share a common approach of using the policy learned in the simplified model as a high-level motion planner, which a full-body controller then tracks. While effective, this strictly hierarchical approach can lead to suboptimal behaviors as the full-body system is constrained by the simplified plan.

In our work, we take this concept a step further by introducing a model continuum 
that bridges the SRB and full-body models.
This approach facilitates a smooth policy transition while enabling the discovery of more optimal behaviors beyond the initial SRB guidance.


\subsection{Continuation Method}
Continuation methods \cite{NumericalContinuation2012}, originating in numerical analysis and optimization, are widely used to address challenging problems by gradually transforming a simpler, well-understood system into a complex target system. This approach is particularly effective for mitigating dependency on initial conditions and avoiding poor local minima. 

The effectiveness of this approach was demonstrated in several recent studies.
For instance, Pardis et al. \cite{ProbabilisticHomotopicOptimization2024} tackled challenging trajectory optimization problems through homotopic optimization. In this method, homotopic paths were explored probabilistically using a tree structure in a multidimensional homotopy parameter space.
Raff et al. \cite{GeneratingGaitFamily2022} introduced a model homotopy to transition between an energetically conservative model (ECM) and a fully actuated robot. Their approach employs parametric scaling to map realistic robot models to their corresponding non-dissipative ECMs. By tracing the passive gaits of the ECM through numerical continuation, they identify optimal actuated gaits for the target system.

Inspired by these approaches, we construct a continuous transition in the model space by defining a $\lambda$-parameterized path that gradually redistributes mass and inertia between the SRB and full-body models.
Although the formal topological conditions of a homotopy are not strictly satisfied due to the hybrid nature of contact dynamics, we use the term 'model homotopic' in the remainder of this paper for brevity to denote this model-homotopy-inspired transition. Furthermore, while our approach shares the broader philosophy of curriculum learning, the latter typically encompasses a wide range of strategies ranging from data scheduling to the escalation of task difficulty.
We specifically adopt the term "Model Homotopy" to more accurately capture the unique nature of our work: the systematic bridging of the model gap by performing continuation along a parameterized dynamics environment.

\begin{figure*}[t]
    \includegraphics[width=0.95\textwidth]{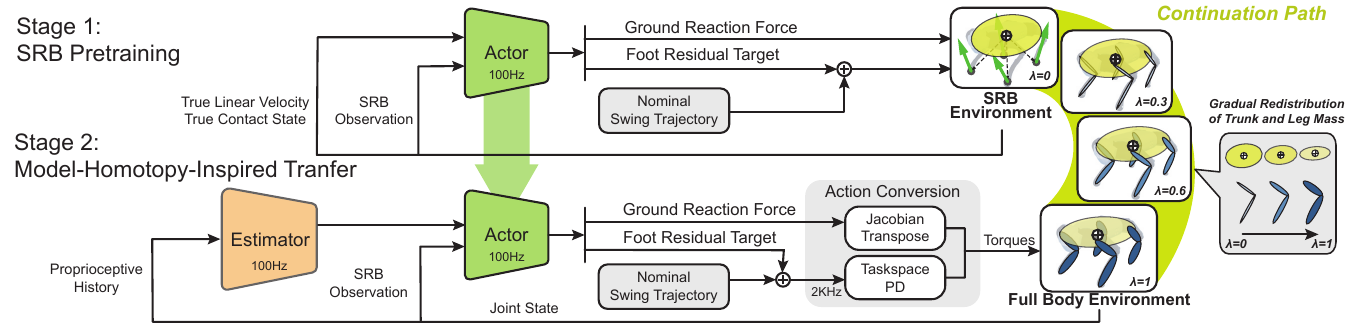}
    \captionsetup{font=small}
    \caption{
    Overview of the proposed learning framework. 
    The motion policy is first pretrained on a simplified SRB model to learn core motion patterns. Subsequently, this policy is smoothly adapted to the full-body environment using our Model-Homotopy-Inspired Transfer.
    }
    \label{figure:Control Framework}
    \vspace{-0.2cm}
\end{figure*}

\section{Simplified Model Pretraining}
In this section, we present SRB pretraining, the first stage of our learning framework.
The goal of this stage is to establish a robust foundation for dynamic behaviors by learning to coordinate body momentum and contact forces under idealized conditions.
By approximating the SRB model as the limiting case where the continuation parameter $\lambda \to 0$ (Sec. IV),
the policy learned during this stage provides a high-quality initial solution for the subsequent transfer to full-body dynamics.

\subsection{Single Rigid Body Model}
\label{ssec:Single Rigid Body Model}
The SRB model represents the robot as a floating rigid body with a fixed lumped mass and inertia, assuming massless legs and ideal point contacts with the ground.
The state comprises the base position, orientation, linear and angular velocities, and the Cartesian positions of the virtual feet.
In this formulation, the ground reaction forces (GRFs) exerted at the contact points serve as the control inputs. The equations of motion for the SRB model, expressed in the world frame, are given as follows:
\begin{align}
    \label{eq:SRB dynamics}
    &\sum_i \mathbf{f}_i + m\mathbf{g} = m \mathbf{\dot v}, 
    &\mathbf{\dot p}&= \mathbf{v} \nonumber \\
    &\sum_i \mathbf{r}_i \times \mathbf{f}_i = \mathbf{I}\mathbf{\dot w} + \mathbf{w} \times \mathbf{I}\mathbf{w}, 
    &\mathbf{\dot R} &= [\mathbf{w}]_{\times} \mathbf{R}
\end{align}
where $\mathbf{r_i}\in \mathbb{R}^3 $ is the vector from the center of mass (COM) to the $i$th contact point, and $\mathbf{f_i}\in \mathbb{R}^3$ represents the GRFs exerted at the $i$th contact point. $\mathbf{p} = [p_x, p_y, p_z]\in \mathbb{R}^3 $ and $\mathbf{v}\in \mathbb{R}^3 $ denote the position and velocity of the COM, respectively. 
$\mathbf{w} \in \mathbb{R}^3$ is the angular velocity,
and $\mathbf{R}= [\mathbf{R_x}, \mathbf{R_y}, \mathbf{R_z}] \in SO(3)$ is a rotation matrix describing the orientation of the body.
$m$ is the lumped mass and $\mathbf{I} = \mathbf{R}\mathbf{I}_{b}\mathbf{R}^\top \in \mathbb{R}^{3\times3}$ is the inertia tensor expressed in the world frame, where $\mathbf{I}_{b}$ denotes the constant inertia matrix in the body-fixed frame.

By effectively capturing essential body dynamics while abstracting away less critical aspects, the SRB model strikes an ideal balance between simplicity and expressiveness. This makes it a suitable middle ground for discovering diverse dynamic behaviors, 
serving as an effective foundation for the initial stage of our learning framework.

\subsection{SRB Pretraining} \label{ssec:SRB_pretraining}
Building upon the dynamics described in the previous section, we create an idealized environment free from confounding factors like leg inertia, leg collisions, and contact instabilities. This simplification allows the policy to focus exclusively on learning the core principles of body momentum control, enabling the efficient discovery of the motion.

\subsubsection{SRB Environment} \label{sssec:SRB_environment}
The SRB model is initialized with the composite mass and inertia of the full robot in its nominal standing pose. Based on this physical configuration, the policy network generates hybrid control actions that alternate between two phases according to a predefined contact plan.
During the stance phase, the policy outputs GRF actions $\mathbf{f_i}$, which are clipped to remain within the friction cone.
Conversely, during the swing phase, the policy outputs a residual foot target actions $\Delta\mathbf{r_i}$. 
These residuals are added to a point on the predefined nominal swing trajectory $\overline{\mathbf{r}}_i$ to determine the final foot target $\hat{\mathbf{r}}_i$. This target is then constrained by clipping to stay within the leg's workspace (a sphere centered at the hip) and to avoid ground penetration. 
Foot contact is determined purely based on kinematic conditions: a foot is considered in contact if the distance between its center and the ground is less than the foot's radius.
Once all control inputs are set, the base's state transition follows the rigid body dynamics in Eq.~\ref{eq:SRB dynamics}.

\begin{table}[t]
\centering
\small
\captionsetup{font=small}
\caption{Contact plan parameters for flip motions.}
\vspace{-0.05cm}  
\label{tab:flip_params}
\begin{tabular}{ll}
\hline
\textbf{Motion} & \textbf{Contact Plan (Air Phase)} \\ \hline
\textbf{Backflip} & Front Legs: [0.15, 0.7]s; Rear Legs: [0.30, 0.85]s \\
\textbf{Sideflip} & Right Legs: [0.15, 0.7]s; Left Legs: [0.30, 0.85]s \\
\textbf{Yawspin} & All Legs: [0.5, 0.9]s \\ \hline
\end{tabular}
\vspace{-0.2cm}  
\end{table}

\subsubsection{Task-Specific Design}
Our framework learns diverse motions, each defined by a high-level structure consisting of a contact plan and a nominal swing trajectory. The contact plan specifies the motion's total duration and the allowed contact intervals for each foot. The nominal swing trajectory defines a central path for the foot during the swing phase, parameterized using a simple quadratic Bezier curve. The policy then learns to output a residual to this nominal path. We designed three classes of motions: gaits, flips, and wall-assisted maneuvers.

Gait motions are defined by periodic stance and swing phases, each with a duration of 0.2~s. The nominal swing trajectory is defined by three control points: the current foot position in the local frame, a target position determined by the Raibert heuristic, and a midpoint with a clearance of 0.1~m.

Flips are composed of takeoff, air, and landing phases over a total duration of 2~s. Key air phase timings are summarized in Table~\ref{tab:flip_params}. The nominal swing trajectory is parameterized as a Bezier curve with three control points: the current foot position, $(0, 0, -0.2)$~m and $(0, 0, -0.3)$~m in the local frame.
\begin{figure}[h]
    \centering
    \small
    \includegraphics[width=0.62\linewidth]{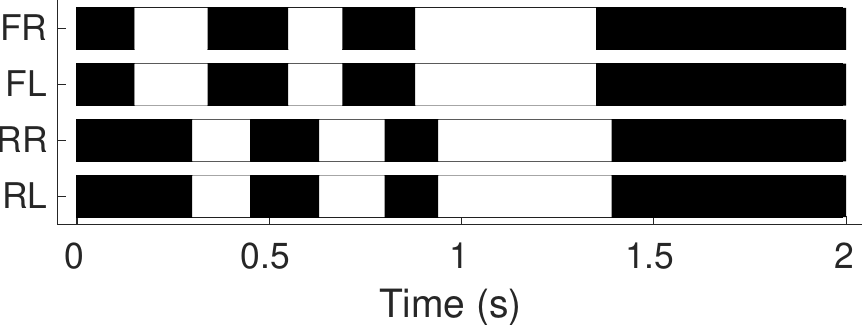}
    \captionsetup{font=small}
    \caption{Contact plan of wall-assisted motions.}
    \label{fig:contactPlan}
    \vspace{-0.2cm}  
\end{figure}

Wall-assisted motions are designed as a series of bounding gaits over a 2~s duration. Following the contact plan illustrated in Fig.~\ref{fig:contactPlan}, each motion is guided by keyframe position and orientation targets that vary across phases. While the targets for the wall-assisted turn are detailed in Table~\ref{tab:wall_assisted_motion_targets}, other maneuvers can be derived by making minor adjustments to these values.
To effectively explore the interaction with the wall, the swing trajectory interpolates between a local path and a world-frame target: $ \overline{\mathbf{r}}_i = \eta\mathbf{r}_b(\phi) + (1-\eta)\mathbf{r}_w(\phi),\ \eta =  16 (\phi-0.5)^4 $
where $\phi \in [0,1]$ represents the motion phase. The local path component, $\mathbf{r}_b$, is defined by a Bezier curve with control points $(-0.1,0,-0.2)$, $(0,0,-0.1)$ and $(0.1,0,-0.2)$. The world-frame target, $\mathbf{r}_w$, is the current foot position if $\phi < 0.5$ and the nominal foot position associated with the upcoming keyframe target otherwise.

\begin{table}[t]
\centering
\captionsetup{font=small}
\caption{Keyframe targets of the wall-assisted turn.}
\vspace{-0.05cm}  
\small
\begin{tabular}{ccc}
    \hline
    \textbf{Phase} & $\mathbf{p}_{\text{target}}$ & $\mathbf{R}_{\text{target}}$ \\ \hline
    Bound (0) & $(0.4, 0, 0.3)$ & - \\
    Air (1) & $(0.4, 0, 0.3)$ & - \\
    Jump Up (2) & $(0.85, 0, 0.6)$ & - \\
    Air (3) & $(0.85, 0, 0.6)$ & $\mathbf{R}_{z}=-\hat{\mathbf{x}}$ \\
    Wall (4) & $(0.2, 0, 0.3)$ & $\mathbf{R}_{z}=-\hat{\mathbf{x}}$ \\
    Air (5) & $(0.2, 0, 0.3)$ & $\mathbf{R}_{z}=\hat{\mathbf{z}}, \mathbf{R}_{x}=-\hat{\mathbf{x}}$ \\
    Landing (6) & $(0, 0, 0.3)$ & $\mathbf{R}_{z}=\hat{\mathbf{z}}, \mathbf{R}_{x}=-\hat{\mathbf{x}}$ \\
    \hline
\end{tabular}
\label{tab:wall_assisted_motion_targets} 
\vspace{-0.2cm}  
\end{table}

\subsubsection{Reward Function}
The reward terms $r_p$, $r_o$, $r_v$, and $r_w$ encourage the body to track the target position, orientation, linear velocity, and angular velocity, respectively. The terms $r_{grf}$ and $r_{res}$ regularize excessive actions. Additionally, $r_{rot}$ is included for motions involving significant rotational maneuvers.
Most reward terms adopt an exponential form defined as $r(\mathbf{x};a,b) = a\exp(-b||\mathbf{x}||^2)$, whereas the rotational reward $r_{rot}$ follows a linear form $r(x;a,u,l) = a \max(\min(x,u),l)$.
The parameters for each motion's reward terms are summarized in Table \ref{tab:reward table}. All notation is consistent with Eq. \ref{eq:SRB dynamics}, unless specified. Reward components active only during specific phases are explicitly marked in the table.
The total reward is calculated as $ ( 1 + \sum r_{\text{pos}} ) \times (\sum r_{\text{neg}})$ where $r_{\text{pos}}$ and $r_{\text{neg}}$ are the positive and negative reward components, respectively.

\begin{table*}[t!]

\centering

\begin{subtable}{\textwidth}
    \centering
    \begin{minipage}{0.28\linewidth}
    \centering
    {\scriptsize
    \begin{threeparttable}
    \begin{tabular}{|c|c|c|c|}
    \hline
    \multicolumn{4}{|c|}{\textbf{Gaits}} \\
    \hline
    \noalign{\vskip 1pt}
    \hline
    negative reward & $\mathbf{x}$ & $a$ & $b$ \\
    \hline
    $r_\textrm{p}$ & $p_\textrm{z}$ - 0.28 & 0.35 & 5 \\
    $r_\textrm{o}$ & $\mathbf{R}_\textrm{z} - \hat{\mathbf{z}}$ & 0.40 & 5 \\
    $r_\textrm{grf}$ & $\mathbf{f}_\textrm{i}$ & 0.05 & 1e-5 \\
    $r_\textrm{res}$ & $\Delta\mathbf{r}_\textrm{i}$ & 0.2 & 2 \\
    \hline
    positive reward & $\mathbf{x}$ & $a$ & $b$ \\
    \hline
    $r_\textrm{v}$ & $\mathbf{v}-\mathbf{v}_\textrm{cmd}$ & 0.3 & 3 \\
    $r_\textrm{w}$ & $\mathbf{w}-\mathbf{w}_\textrm{cmd}$ & 0.3 & 3 \\
    \hline
    \end{tabular}
    \begin{tablenotes}        
        \item $\mathbf{v}_\textrm{cmd}$ and $\mathbf{w}_\textrm{cmd}$ denote the command linear and angular velocities, respectively.
    \end{tablenotes}
    \end{threeparttable}
    }
    \end{minipage}
\hfill
    \begin{minipage}{0.33\linewidth}
    \centering
    {\scriptsize
    \begin{threeparttable}
    \begin{tabular}{|c|c|c|c|}
    \hline
    \multicolumn{4}{|c|}{\textbf{Flips}} \\
    \hline
    \noalign{\vskip 1pt}
    \hline
    negative reward & $\mathbf{x}$ & $a$ & $b$ \\
    \hline
    $r_\textrm{p}$ & $\mathbf{p}-0.3 \hat{\mathbf{z}}$& 0.1 & 1 \\
    $r_\textrm{o}$ (landing) & $\mathbf{R}_\textrm{z}-\hat{\mathbf{z}}$ & 0.15 & 0.1 \\
    $r_\textrm{w}$ & $ \mathbf{w} - (\mathbf{\alpha}^t \mathbf{w})\mathbf{\alpha}$ & 0.2 & 1 \\
    $r_\textrm{grf}$ & $\mathbf{f}_\textrm{i}$ & 0.05 & 1e-5 \\
    $r_\textrm{res}$ & $\Delta\mathbf{r}_\textrm{i}$ & 0.2 & 2 \\
    \hline
    positive reward & $\mathbf{x}$ & $a$ & $(u, l)$ \\
    \hline
    $r_\textrm{rot}$ (air) & $\mathbf{\alpha}^t \mathbf{w}$ & 0.3 & (6, -0.1) \\
    \hline
    \end{tabular}
    \begin{tablenotes}       
        \item $\mathbf{\alpha}$ is the desired rotational axis, which is  $\hat{\mathbf{x}}$ for sideflip, $\hat{\mathbf{y}}$ for backflip and $\hat{\mathbf{z}}$ for yawspin.
    \end{tablenotes}
    \end{threeparttable}
    }
    \end{minipage}
\hfill
    \begin{minipage}{0.33\linewidth}
    \centering
    {\scriptsize
    \begin{threeparttable}
    \begin{tabular}{|c|c|c|c|}
    \hline
    \multicolumn{4}{|c|}{\textbf{Wall Assisted Motions}} \\
    \hline
    \hline
    negative reward & $\mathbf{x}$ & $a$ & $b$ \\
    \hline
    $r_\textrm{p}$ & $\mathbf{p}-\mathbf{p}_{\textrm{target}}$ & 2 & 0.1 \\
    $r_\textrm{o}$ & $\mathbf{R}-\mathbf{R}_{\textrm{target}}$ & 1 & 0.1 \\
    $r_\textrm{w}$ & $\mathbf{w} - (\mathbf{\alpha}^t \mathbf{w})\mathbf{\alpha}$ & 0.15 & 0.01 \\
    $r_\textrm{grf}$ & $\mathbf{f}_\textrm{i}$ & 0.05 & 1e-5 \\
    $r_\textrm{res}$ & $\Delta\mathbf{r}_\textrm{i}$ & 0.2 & 2 \\
    \hline
    positive reward & $\mathbf{x}$ & $a$ & $(u,l)$ \\
    \hline
    $r_\textrm{rot}$ (air) & $ \mathbf{\alpha}^t \mathbf{w}$  & 1 & (8,-0.1) \\
    \hline
    \end{tabular}
    \begin{tablenotes}       
        \item $\mathbf{\alpha}$ is $\hat{\mathbf{y}}$ for wall assisted backflip and $\mathbf{0}$ otherwise.\\
    \end{tablenotes}
    \end{threeparttable}
    }
    \end{minipage}
\end{subtable}
\captionsetup{font=small}
\caption{Reward Table of Generated Motions}
\vspace{-0.3cm}
\label{tab:reward table}
\end{table*}

\subsubsection{Policy Architecture and Training}
The policy is implemented as a Multi-Layer Perceptron (MLP) and trained using Proximal Policy Optimization (PPO). The network's inputs consist of the gravity vector, linear and angular velocities, Cartesian positions of the four feet in the local base frame, contact states, and the sine and cosine of the current phase. 
For linear velocity and contact states, the true values from the simulation are used as privileged information, which are later substituted by the estimator network's predictions in the full-body environment (Sec.\ref{ssec:estimation_network}).
For each learning iteration, data is collected from 100 parallel environments over a 4-second rollout and trained for eight epochs.

\section{Model-Homotopy-Inspired Transfer}  \label{sec:homotopy_continuation_transfer}
In this section, we address the challenge of bridging the significant model gap between the SRB model and the full-body model. 
We adopt a continuation transfer approach inspired by model homotopy, which constructs a continuous path through progressive adjustment of inertial parameters to facilitate a smooth transition.
The overall structure of this transfer process is illustrated in Fig. 2.
\subsection{Model Homotopic Environment} \label{ssec:continuation_environment}
We formulate a continuous environment that smoothly bridges the SRB and full-body model by gradually increasing the leg mass and inertia from nearly zero to their original values. 
To keep the total mass constant and prevent large changes in the GRF action, we also adjusted the trunk mass, inertia, and its center of mass position to linearly interpolate between the composite values from the nominal pose and their original full-body values.
This entire process of redistributing mass between the trunk and legs can be parameterized by a continuation parameter $\lambda \in (0, 1]$, defined as the ratio of leg mass to the original leg mass. 
The inertial properties of the system are parameterized as a function of $\lambda$ as follows:
\begin{align}
    \label{eq:curriculum}
    m_{leg}(\lambda) &= \lambda\cdot m_{leg,full}\nonumber\\
    \mathbf{I}_{leg}(\lambda) &= \lambda\cdot \mathbf{I}_{leg, full} \nonumber\\
    m_{trunk}(\lambda) &= \lambda\cdot m_{trunk, full} + (1-\lambda)\cdot m_{composite}\\
    \mathbf{I}_{trunk}(\lambda) &= \lambda\cdot \mathbf{I}_{trunk, full} + (1-\lambda)\cdot\mathbf{I}_{composite}\nonumber \\
    \mathbf{c}_{trunk}(\lambda) &= \lambda \cdot \mathbf{c}_{trunk, full} + (1-\lambda) \cdot \mathbf{c}_{composite} \nonumber 
\end{align}
where $m_{trunk, full}$, $\mathbf{I}_{trunk, full}$, and $\mathbf{c}_{trunk, full}$ are the original trunk mass, inertia, and CoM, respectively. $m_{composite}$, $\mathbf{I}_{composite}$, and $\mathbf{c}_{composite}$ represent the composite properties of the entire robot in the nominal pose.

This parametrization defines a family of dynamics mappings $H: X \times (0, 1] \to Y$, where $X$ and $Y$ represent the full-body state-action space and resulting accelerations, respectively.
Since the equations of motion depend continuously on the inertial parameters, which vary linearly with $\lambda$, the mapping $H(x, \lambda)$ maintains strict continuity with respect to the continuation parameter. 
Consequently, training within this $\lambda$-parameterized environment can be framed as a numerical continuation process. This allows the policy to incrementally adapt to the evolving dynamics and reward landscape, promoting stable convergence to a desirable local optimum in the target environment.

We employ a piecewise linear schedule to transition $\lambda$ from 0.01 to 1.0 over 900 iterations. 
To balance training efficiency and stability, we adopt a decaying step size: starting with an increment of 0.01 for $\lambda < 0.6$, we reduce it to 0.005 for $\lambda \in [0.01, 0.8)$, and to 0.0025 as $\lambda$ reaches 1.0.
Additionally, to mitigate numerical instability caused by the ill-conditioned mass matrix at low $\lambda$ values, we adaptively adjusted the simulation timestep, starting with finer time steps for small $\lambda$ and gradually increasing them to the nominal value as $\lambda$ increases.

\subsection{Action Conversion}
While the controller maintains the structure described in Sec.~\ref{ssec:SRB_pretraining}, an additional action conversion step is required to translate the actions into actual full-body torques.

During the stance phase, if the foot is in contact, we transform GRFs into joint torques using the Jacobian transpose:
\begin{align}
    \boldsymbol{\tau}_i^{\text{stance}} = \mathbf{J}_i^T \mathbf{f}_i
\end{align}
where $\mathbf{J}_i$ is the $i$th foot Jacobian, and $\mathbf{f}_i, \boldsymbol{\tau}_i^{\text{stance}} \in \mathbb{R}^3$ are the GRF action and resulting torque for $i$th stance leg. 
During the swing phase, task-space PD control is used to track the foot target, with the PD gains set to \(K_p = 500\) and \(K_d = 5\).
\begin{align}
        \hat{\mathbf{r}}_i &=  \overline{\mathbf{r}}_i + \Delta\mathbf{r}_i \nonumber \\
        \boldsymbol{\tau}_i^{\text{swing}} &= \mathbf{J}_i^T  [K_p(\hat{\mathbf{r}}_i-\mathbf{r}_i)+K_d(\hat{\mathbf{s}}_i-\mathbf{s}_i)]
\end{align}
where $\mathbf{r}_i \in \mathbb{R}^3$ and $\mathbf{s}_i \in \mathbb{R}^3$ are the current position and velocity of the $i$th foot, and $\hat{\mathbf{s}}_i$ is set to zero in our case. $\boldsymbol{\tau}_i^{\text{swing}} \in \mathbb{R}^3$ is the torque for $i$th swing leg.

\subsection{Estimation Network}  \label{ssec:estimation_network}
We utilized the concurrent estimation framework\cite{Concurrent2022} to infer linear velocity and contact states from proprioceptive sensor histories, which are subsequently incorporated into the policy's observation. 
Modeled as a two-layer MLP, the estimator is trained concurrently with the policy by minimizing the loss function $\mathcal{L}_{\text{est}} = ||\hat{\mathbf{v}} - \mathbf{v}||^2 + ||\hat{c} - c||^2$, where $\hat{\mathbf{v}}$ and $\hat{c}$ denote the estimated linear velocity and contact probabilities, while $\mathbf{v}$ and $c$ represent the ground truth values from simulation. 
Network inputs comprise the history of joint positions, velocities, and torque commands, concatenated with the non-privileged policy observations.
The discrete contact state is determined by applying a threshold of 0.5 to the estimated probability.

\begin{figure*}[ht!]
    \begin{subfigure}{\linewidth}
        \centering
        \includegraphics[width=0.86\textwidth]{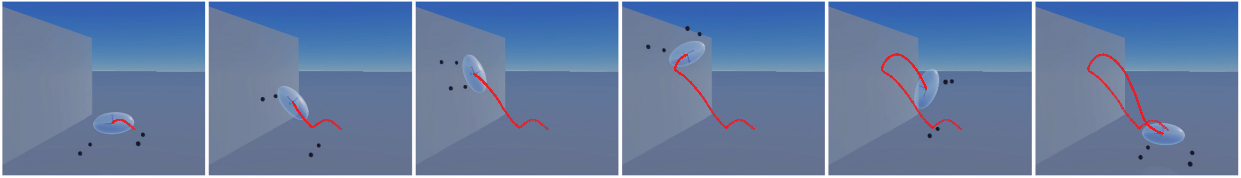}
        \includegraphics[width=0.86\textwidth]{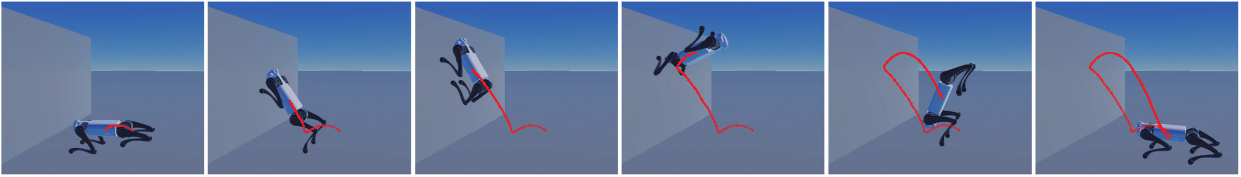}
    \end{subfigure}
    \captionsetup{font=small}
    \caption{Snapshots of wall-assisted backflip. The red lines represent the CoM trajectories over time.}
    \label{figure:wall assisted motions}
    \vspace{-0.2cm}  
\end{figure*}

\section{Results}
\label{sec:Results}
This section details the results in both simulated and real-world settings. We evaluate the effectiveness of Model-Homotopy-Inspired Transfer (MHIT) for seamless motion transfer and provide a comparative analysis against established baseline methods. Additionally, we showcase the successful deployment of the generated motions on real hardware.

\subsection{Motion Generation}
To verify the effectiveness of our method, we evaluated its ability to generate motions challenging to obtain via standard end-to-end RL.
The proposed framework successfully generated a wide range of complex motions in simulation. We produced various locomotion gaits, as well as flipping motions such as backflip, sideflip, and yawspin. Beyond these, we also explored dynamic maneuvers that leverage environmental interactions for rapid redirection and momentum changes. 
The wall-assisted jump example, as shown in Fig.~\ref{figure:model_homotopic_path}, demonstrates the agent using the wall as support for upward propulsion to reach a higher platform. Similarly, the wall-assisted backflip shows the agent initiating a backflip by applying force against the wall to propel itself into the air. 
Fig.~\ref{figure:wall assisted motions} illustrates the wall-assisted backflip, showcasing both the pretrained SRB motion and the corresponding full-body motion after transfer. 
For a comprehensive visualization of these results, please refer to the supplementary video.

\subsection{Analysis of the Proposed Methodology} 

\subsubsection{Effectiveness of Model Homotopy in Motion Transfer}
To investigate the efficacy of MHIT as a strategy for transferring the SRB policy to the full-body environment, we compared the proposed method against several baseline approaches, in terms of sample efficiency, learning stability, and optimality. The baseline methods used are:
\begin{itemize}
    \item \textbf{Direct Transfer (DT):} The SRB policy is directly fine-tuned in the full-body environment without intermediate adaptation steps.
    \item \textbf{Model-Homotopy-Inspired Transfer (MHIT, Ours):} Training progresses along a series of model homotopic environments for the first 900 iterations, followed by additional training in the full-body environment.
    \item \textbf{Imitation Transfer (IT):} This method employs imitation learning with Reference State Initialization (RSI) and Early Termination (ET). A trajectory recorded from the SRB policy served as the kinematic reference, utilizing the same reward settings as \cite{DeepMimic2018}.
    \item \textbf{Vanilla RL:} Trained solely with motion rewards, without pretraining or imitation.
\end{itemize}

\begin{figure}[t]
    \centering
    \begin{minipage}[c]{0.02\columnwidth}
        \small \textbf{(a)}
    \end{minipage}
    \hfill
    \begin{minipage}[c]{0.95\columnwidth}
        \centering
        \includegraphics[width=0.93\linewidth]{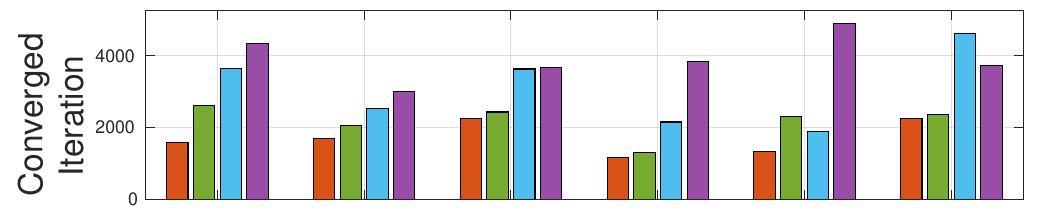}
        \label{figure:convergence_iteration}
    \end{minipage}
    \begin{minipage}[c]{0.02\columnwidth}
        \small \textbf{(b)}
    \end{minipage}
    \hfill
    \begin{minipage}[c]{0.95\columnwidth}
        \centering
        \includegraphics[width=0.93\linewidth]{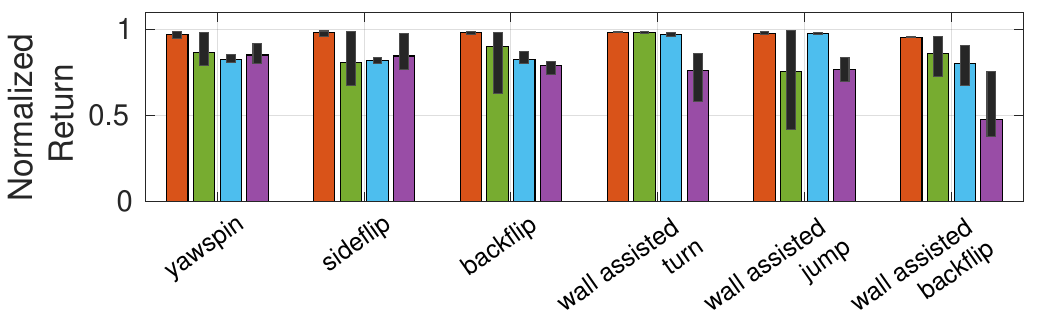}
        \label{figure:normalized_return}
    \end{minipage}
    \begin{minipage}[c]{0.02\columnwidth}
        \small \textbf{(c)}
    \end{minipage}
    \hfill
    \begin{minipage}[c]{0.96\columnwidth}
        \centering
        \includegraphics[width=0.88\linewidth]{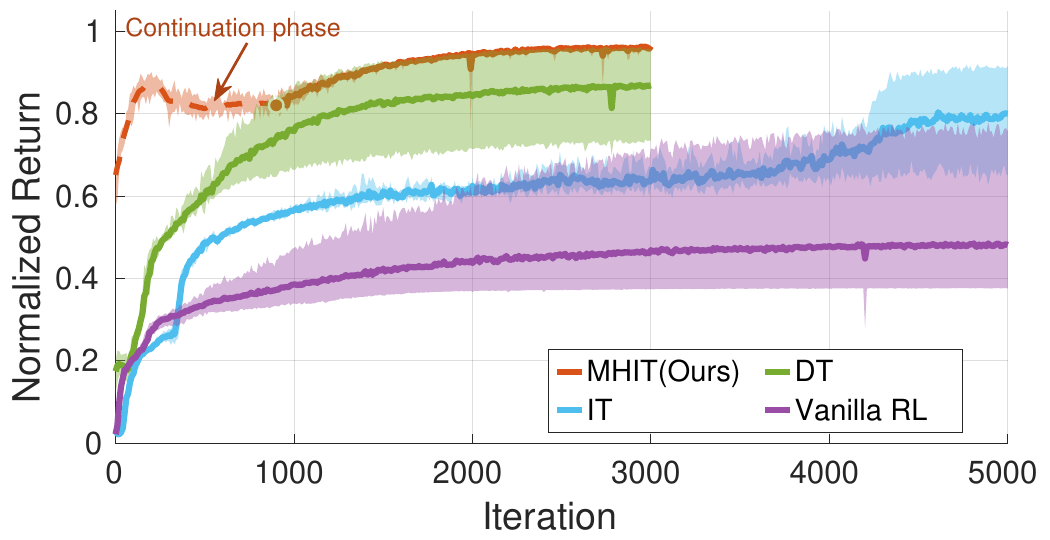}
        \label{figure:learning_curve}
    \end{minipage}
    \captionsetup{font=small}
    \caption{Comparison of training performance across various methods and motions. 
    (a) Converged iteration. The number of iterations required for the return to reach 99\% of the final mean value. 
    (b) Normalized return. The black box on top of each bar indicates the range between the maximum and minimum normalized returns.
    (c) Learning curves for wall-assisted backflip. The solid line represents the average, while the shaded region shows the range between maximum and minimum values.
    The dashed line indicates the continuation phase (the first 900 iterations), while the solid line represents the subsequent training in the target environment.
    Overall, the proposed MHIT demonstrated the fastest convergence and the most consistent performance.}
    \label{figure:analysis}
    \vspace{-0.4cm}  
\end{figure}

The results are presented in Fig.~\ref{figure:analysis}. For each method, training was conducted five times using distinct initial (pretrained or random) policies and seeds. The SRB pretraining is omitted from the figure, as it is computationally inexpensive and serves as a common initialization for all baselines except Vanilla RL.

Overall, our method achieves the fastest convergence, as shown in Fig.~\ref{figure:analysis} (a), and the most consistent learning outcomes, as shown in Fig.~\ref{figure:analysis} (b). 
Specifically, in the wall-assisted backflip task, MHIT achieves a 19.0\% higher normalized return and converges twice as fast as IT (2,240 vs. 4,610 iterations). Additionally, it demonstrates superior stability over DT, maintaining a tight variation (max-min range) of 0.0042 compared to 0.2285 for DT, as shown in Fig.~\ref{figure:analysis} (c).

Beyond normalized returns, we further evaluated the generated motions by comparing several key metrics for the best-performing seeds. MHIT demonstrated superior rotational performance over IT, achieving lower keyframe orientation error ($\|\mathbf{R}-\mathbf{R}_{\textrm{target}}\|$, 0.129 vs. 0.185), higher pitch rotational velocity ($\mathbf{\alpha}^t \mathbf{w}$, 2.004 vs. 1.794 rad/s), and reduced cross-axis angular velocity error ($\|\mathbf{w} - (\mathbf{\alpha}^t \mathbf{w})\mathbf{\alpha}\|$, 0.368 vs. 0.493 rad/s).

The superior performance of our approach can be attributed to two main factors. First, the pretrained policy allowed the agent to retain reactive behaviors learned during pretraining and initiate exploration around promising action spaces. Second, the smooth transition facilitated by the model homotopic environment enabled the policy to adapt to the full-body model while operating in regions of high success rate, continuously generating high-quality samples. This significantly improved learning stability and sample efficiency.

In contrast, DT suffered from significant performance degradation early in training due to the model gap, hindering learning by accumulating poor-quality samples. This occasionally resulted in the loss of behaviors learned during pretraining. 
The IT approach, while benefiting from dense references derived from SRB trajectories and showing relatively stable training compared to DT, often converged to suboptimal policies because the model gap between the SRB reference and the full-body environment persisted.  Furthermore, starting from a random initial policy required the policy to re-discover the motion, which significantly slowed convergence. 
Conversely, the Vanilla RL, trained entirely from scratch without pretraining, consistently failed to discover meaningful motions. This highlights the effectiveness of SRB pretraining, which succeeds in identifying essential motion patterns.

\subsubsection{Comparison of Disturbance Robustness}
To further evaluate the performance of the learned policies, we conducted disturbance robustness tests on MHIT and IT method. 
The success rate was measured across extensive combinations of force and torque disturbances. 
For each combination, 100 different disturbance directions were tested at the specified magnitude, and the average success rate was recorded.
Disturbances were applied continuously throughout the trajectory. Note that no disturbances were experienced during training.
 \begin{figure}[t]
        \centering
        \includegraphics[width=0.90\columnwidth]{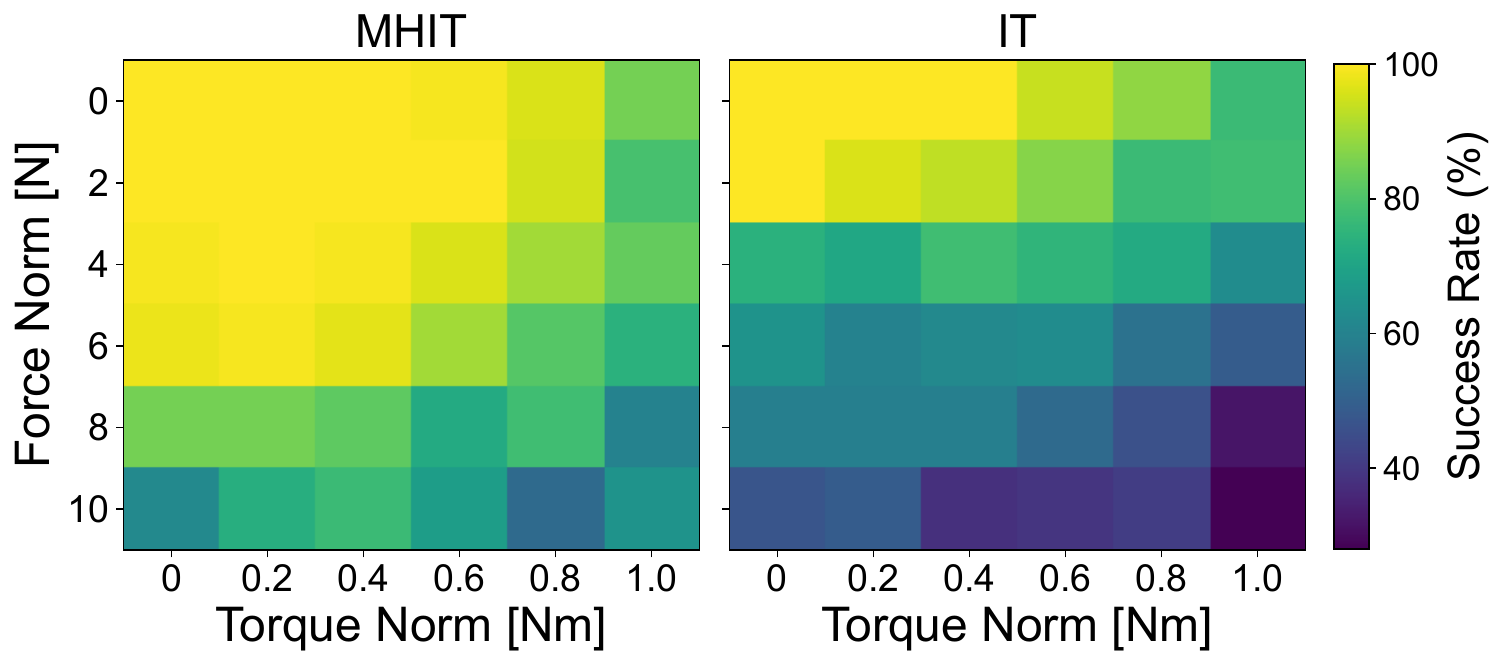}
    \captionsetup{font=small, skip=3pt}
    \caption{Comparison of robustness across disturbances. Each heatmap shows the success rate for combinations of force and torque disturbance norms, with brighter colors indicating higher success rates.}
    \label{figure:disturbance robustness}
    \vspace{-0.2cm}  
\end{figure}

The results, illustrated in Fig.~\ref{figure:disturbance robustness}, show that the MHIT exhibited greater robustness compared to IT. 
This enhanced robustness is likely due to the capability of MHIT to preserve reactive behaviors acquired during pretraining. Moreover, by exploring the state space without the constraints of ET, the policy can experience a wider range of deviations and learn to stabilize from diverse states.



\subsection{Real-World Deployment}
We validated our method by deploying two representative policies, a trot gait and a backflip, on a Unitree Go1 robot.
To increase the fidelity of the simulation to the physical world, we incorporated rotor inertia, sensor and motor command latency, and realistic actuator characteristics into the full-body simulation.
Robustness was further enhanced through domain randomization applied to initial poses, friction coefficients, trunk inertia, motor strength, and sensor noise injection.

In the trot gait experiments, we evaluated velocity tracking performance for a forward velocity of $1.0$ m/s and a yaw rate of $0.3$ rad/s.
We compared our method against an SRB-pretrained policy directly deployed without any fine-tuning.
While both policies were able to generate a stable trot gait without falling, the SRB policy exhibited significant steady-state errors due to the inherent model mismatch, as quantified in Fig.~\ref{figure:combined_results} (a).

\begin{figure}[t]
    \centering
    \includegraphics[width=0.97\columnwidth]{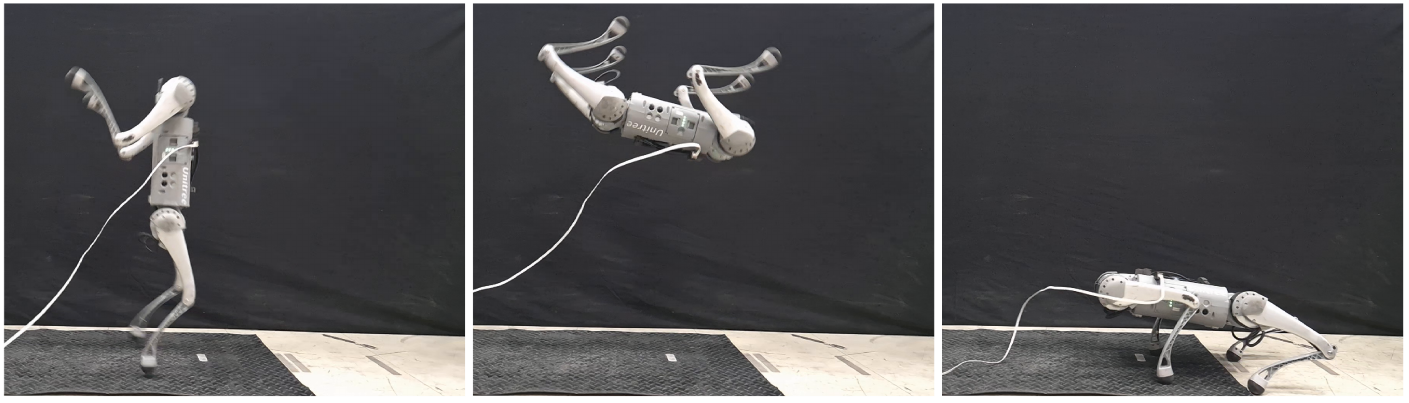}
    \captionsetup{font=small}
    \caption{Snapshots of the backflip execution on the Unitree Go1 robot.}
    \label{figure:real_experiment}
    \vspace{-0.2cm}
\end{figure}

\begin{figure}[t!]
    \centering
    \begin{minipage}[c]{0.02\columnwidth}
        \footnotesize \textbf{(a)}
    \end{minipage}
    \hfill
    \begin{minipage}[c]{0.96\columnwidth}
        \centering
        \includegraphics[width=0.49\linewidth]{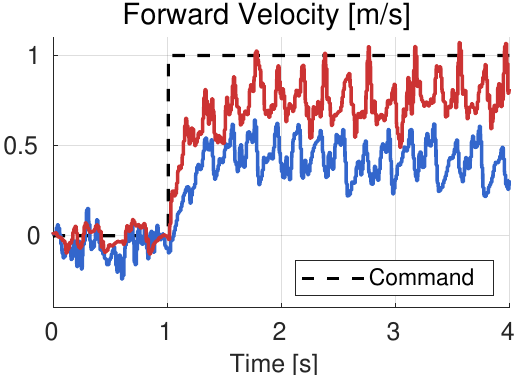}
        \includegraphics[width=0.49\linewidth]{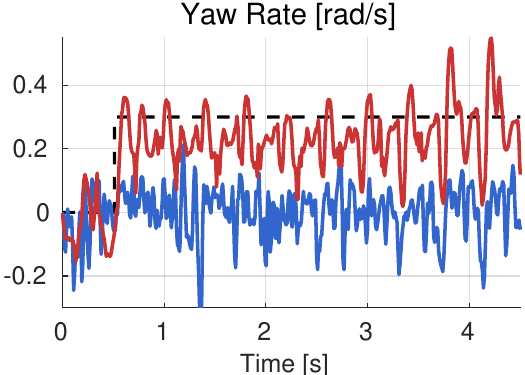}
    \end{minipage}
    

    \begin{minipage}[c]{0.02\columnwidth}
        \footnotesize \textbf{(b)}
    \end{minipage}
    \hfill
    \begin{minipage}[c]{0.96\columnwidth}
        \centering
        \includegraphics[width=0.49\linewidth]{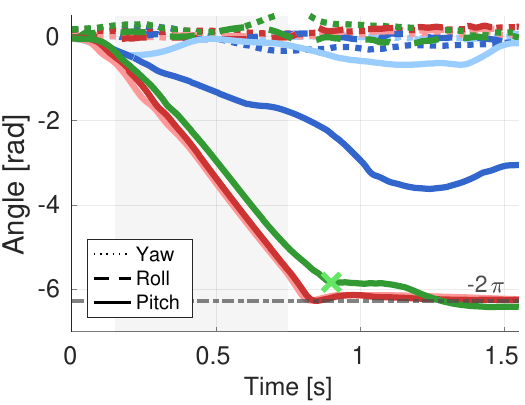}
        \includegraphics[width=0.49\linewidth]{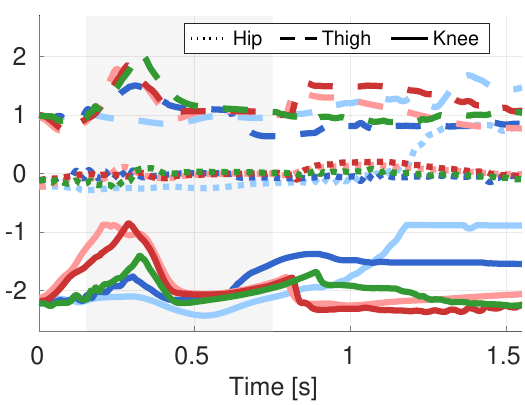}

    \end{minipage}
    \includegraphics[width=0.95\columnwidth]{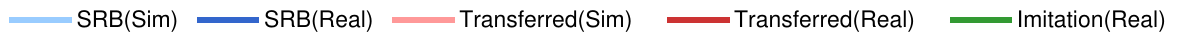}

    \captionsetup{font=small}
    \caption{Quantitative analysis of real-world deployment. 
    (a) Trot velocity tracking: Comparison between the SRB and transferred policies. The SRB policy resulted in a velocity tracking RMSE of 0.53 m/s and a yaw rate RMSE of 0.29 rad/s, whereas the transferred policy achieved 0.28 m/s and 0.11 rad/s, respectively, demonstrating superior tracking accuracy.
    (b) Backflip trajectories: Base orientation (left) and right-rear leg joint angles (right) over time. The trajectories of the IT baseline are included for reference, where the 'x' symbol denotes the instance of frontal impact before reaching a landing configuration.
    }
    \label{figure:combined_results}
    \vspace{-0.2cm}
\end{figure}

The benefits of MHIT were particularly evident in the challenging backflip motion. 
Although successful in SRB simulation, the SRB policy failed in the real world as it could not account for the increased composite inertia during leg extension. This resulted in insufficient angular momentum, causing severe under-rotation and leading the robot to land on its back.
In contrast, the policy transferred using our method successfully executed the backflip on the physical robot, as shown in Fig.~\ref{figure:real_experiment}, achieving close alignment between simulated and real-world trajectories, as illustrated in Fig.~\ref{figure:combined_results} (b).
Furthermore, the policy achieved a 90\% success rate over 10 consecutive trials, demonstrating the practical robustness of the proposed method.
For a visual demonstration of these real-world deployments, please refer to the supplementary video.

\section{Conclusion}
In this study, we introduced a framework for complex dynamic motions using simplified model pretraining and Model-Homotopy-Inspired Transfer to bridge the model gap. This approach enables efficient discovery of core motion patterns and facilitates a smooth transition to full-body dynamics. Validated through maneuvers like backflips and wall-assisted motions, our method outperformed imitation-based baselines in both convergence and robustness, with successful real-world deployment highlighting its practical applicability.

Despite these results, the framework's reliance on predefined contact sequences remains a limitation. 
While foot residuals provide adaptability to mildly uneven ground, fixed temporal structures can lead to instability on significant terrain variations where a change in contact timing is required. Future work will focus on automating contact schedules to enhance environmental adaptability.
Furthermore, expanding this framework to diverse morphologies is a promising direction. 
By abstracting the action space into morphology-independent commands, such as base accelerations, a framework capable of providing a shared evolutionary starting point for a wide range of floating-base systems would be an interesting future direction.

\bibliography{Format/refs} 

\bibliographystyle{IEEEtran}

\end{document}